\documentclass[10pt,twocolumn,letterpaper]{article}

\usepackage[pagenumbers]{cvpr} 

\usepackage{cuted}
\usepackage{fontawesome5}

\usepackage{algorithm}
\usepackage{algorithmic}
\usepackage{multirow}
\usepackage{booktabs}
\usepackage{amssymb}
\usepackage{amsmath}
\usepackage[table]{xcolor}
\usepackage{makecell}
\usepackage{colortbl}    
\usepackage{enumitem}
\usepackage{gradient-text}
\usepackage{xspace}
\usepackage{gradient-text}

\definecolor{cvprblue}{rgb}{0.21,0.49,0.74}
\usepackage[pagebackref,breaklinks,colorlinks,allcolors=cvprblue]{hyperref}

\newcommand{\ours}{\gradientRGB{WaveZip}{4,120,87}{14,165,233}\xspace}

\def\paperID{*****} 
\def\confName{CVPR}
\def\confYear{2026}

\title{\ours: Wavelet-Driven Space-Time Decoupling for \\ Video Token Condensation}

\author{
Yuhui Zeng$^1$, Wang Chen$^1$, Jinfa Huang$^2$, Tianyu Xie$^1$, 
Yongdong Luo$^1$, \\ Jiayi Ji$^1$, Xiawu Zheng$^1$, jiebo Luo$^2$  \\
$^1$Media Analytics and Computing Lab, Xiamen University, Xiamen, China \\ $^2$Department of Computer Science, University of Rochester, Rochester, NY, USA
}

\begin{document}
\maketitle

\begin{abstract}
  Existing Large Vision-Language Models (LVLMs) struggle with long-form video understanding due to the quadratic computational cost of visual tokens. 
  While recent efficient methods attempt to compress tokens via hard pruning or uniform merging, they operate strictly in the spatial feature domain, where robust structural context and discriminative semantic details are inherently entangled. 
  In this work, we propose \textbf{\ours}, a joint signal-frequency-domain framework for efficient video inference. 
  Driven by the insight that temporal redundancy resides in low-pass approximation scales while spatial saliency strongly correlates with high-frequency components, \ours\ leverages Discrete Wavelet Transforms (DWT) to disentangle these signals. 
  Temporally, it employs 1D DWT to analyze query-frame relevance, and the resulting high-frequency coefficients are further gated by inter-frame differences, with both signals jointly driving the dynamic allocation of a precise frame-level token budget.
  Spatially, a 2D DWT decomposes features into low-frequency approximations and high-frequency detail components, where the high-frequency coefficients are modulated within query-salient regions to regulate spatial reconstruction.
  Importantly, \ours\ requires no task-specific training and can be seamlessly integrated into off-the-shelf LVLMs to boost inference efficiency. 
  Extensive experiments on long video understanding benchmarks demonstrate that \ours\ retains 99.6\% of the full performance 
  under an extreme 10$\times$ compression ratio, consistently outperforming state-of-the-art methods.
\end{abstract}    
\section{Introduction}
\label{sec:intro}
Large Vision-Language Models (LVLMs)~\cite{bai2025qwen2, zhang2024video, zhang2023video, wang2025internvl3, li2023videochat, hurst2024gpt} have rapidly advanced video understanding, enabling richer reasoning over complex, long-form visual content.
Despite this strong potential, the transition from short clips to long videos exposes a severe computational bottleneck.
Long videos produce thousands of visual tokens across space and time, causing the transformer attention to scale quadratically with token length.
Consequently, efficient token condensation has become essential for practical LVLM inference.


\begin{figure}[t]
    \centering
    \includegraphics[width=1\linewidth]{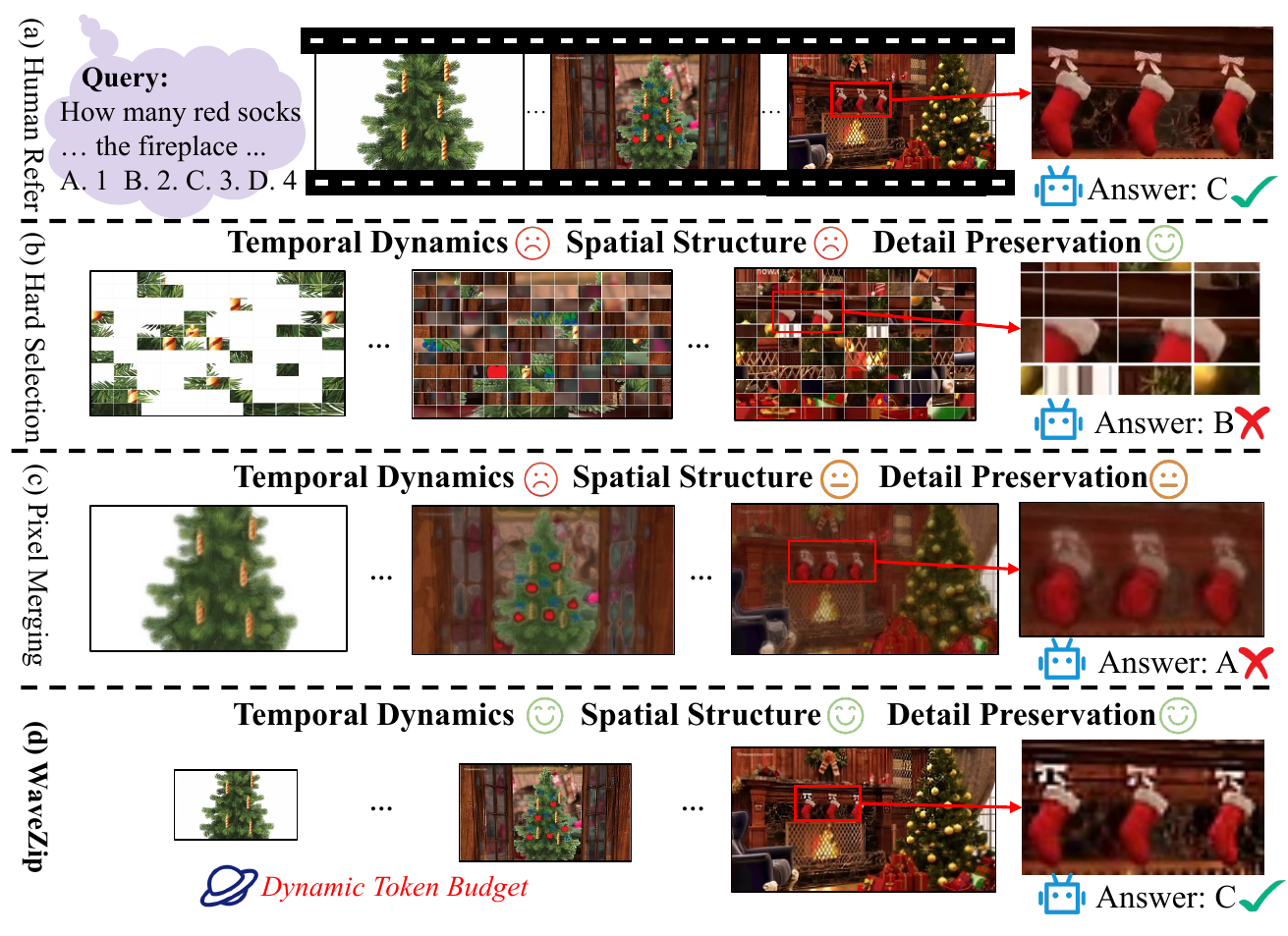}
    \caption{\textbf{Comparison of token compression strategies.}
    Unlike hard selection or uniform merging, WaveZip performs query-conditioned token condensation jointly across time and space, balancing temporal dynamics, spatial structure, and fine-grained details.}
    \label{fig:method_comp}
\end{figure}
Existing approaches attempt to alleviate this bottleneck by compressing visual tokens directly in the spatial feature domain, typically via hard selection or uniform merging.
%
%
Hard selection applies binary keep-or-discard decisions that can disrupt spatial coherence and remove structured visual context.
Uniform merging applies a fixed aggregation operator that behaves like a global low-pass filter, inevitably blurring semantically salient regions.


We view a key limitation of direct feature-domain compression as \emph{frequency conflation}.
Temporally, frame-query relevance sequences mix slowly varying semantic trends with local high-frequency fluctuations caused by sampling noise, visual jitter, or meaningful scene changes.
Treating these fluctuations uniformly can make frame-level
budget allocation unstable.
Spatially, task-relevant fine-grained evidence and redundant background texture remain mixed within the same feature maps, making it difficult to preserve the former without retaining the latter.
This motivates a frequency-domain formulation with two complementary components: temporal relevance decomposition and spatial feature decomposition. 
The former separates frame-level relevance into trend and detail signals, with the detail signal conditioned on inter-frame visual change. 
The latter separates frame features into low-frequency structure and high-frequency detail, with query-conditioned guidance applied selectively to the detail bands.
Our frequency-band analysis further shows that query-conditioned saliency is more strongly correlated with the wavelet detail bands than with the low-frequency approximation band.

Based on this perspective, we propose \ours, a wavelet-driven framework for spatiotemporal token condensation. 
\ours uses DWT because it provides a localized, invertible, and multi-band decomposition. 
Temporally, we apply a 1D DWT to the frame-level relevance sequence and use an inter-frame visual-change prior to gate its detail coefficients. 
Inverse DWT then reconstructs a rectified relevance sequence for adaptive frame-level budget allocation. 
Spatially, we apply a 2D DWT to each frame feature map, retain the low-frequency approximation, and reweight the high-frequency detail subbands using query-conditioned spatial saliency. 
The reconstructed frame features are finally downsampled according to the allocated budgets and projected into compact visual tokens. 
The entire pipeline requires no task-specific optimization and can be integrated into off-the-shelf LVLMs.

Our contributions are summarized as follows:
\begin{itemize}
    \item We highlight frequency conflation as a key limitation of direct feature-domain token compression and formulate spatiotemporal token condensation as a frequency-aware decoupling problem.
    
    \item We propose \ours, a plug-and-play framework requiring no task-specific training. 
    \ours rectifies temporal relevance through visual-change-gated wavelet decomposition and preserves query-relevant spatial details through saliency-guided high-frequency reweighting.

    \item Extensive experiments across multiple long-video benchmarks and LVLM backbones show that \ours consistently improves the accuracy--compression trade-off, retaining \textbf{99.6\%} of the full-token performance under \textbf{$10\times$} compression.

\end{itemize}

\section{Related Work}
\label{sec:related_work}

\subsection{LVLMs and Visual Token Compression}

Large Vision-Language Models (LVLMs) have shown strong video understanding capabilities by coupling visual encoders with LLMs for video question answering, temporal reasoning, and multimodal dialogue~\cite{zhang2024video, lin2024video, li2024llava, liu2023visual, bai2025qwen2, li2023blip, hurst2024gpt, zhang2023video, li2023videochat, li2024llama, liu2024st}. 
However, long-form videos introduce a large number of visual tokens, and the quadratic attention cost of LLM prefilling makes efficient inference increasingly difficult. 
This has motivated visual token compression methods that reduce redundant tokens before or during LVLM inference.

Existing compression methods mainly follow two paradigms. 
Hard selection methods remove visually less important tokens using attention, similarity, or learned scoring signals~\cite{chen2024image, xing2024pyramiddrop, yang2025visionzip, yang2025vflowopt, shang2025llava, zeng2025glimpse, liu2025hiprune, zhang2025beyond, luo2026quota}. 
Representative methods such as FastV~\cite{chen2024image} and VisionZip~\cite{yang2025visionzip} achieve efficient pruning, but the binary keep-or-discard operation can break spatial coherence and remove structured visual context. 
Merge-based methods instead aggregate redundant tokens into compact representations~\cite{bolya2022token, shen2025fastvid, shen2024tempme, shao2025holitom, tao2025dycoke, huang2025tosa, wan-etal-2024-look}. 
For example, FastVID~\cite{shen2025fastvid}, DyCoke~\cite{tao2025dycoke}, and PruMerge~\cite{shang2025llava} reduce token redundancy through dynamic pruning or merging, but their operations are still performed directly in token or feature space.
Consequently, existing methods generally lack explicit control over low-frequency structural trends and high-frequency task-relevant details, motivating our frequency-domain formulation.

\subsection{Frequency-Domain Compression}
Frequency-domain analysis provides a natural way to separate coarse structure from fine-grained details. 
Wave-ViT~\cite{yao2022wave} introduces wavelet decomposition into vision transformers to preserve high-frequency texture during multi-scale representation learning, but it is not designed for query-conditioned video token condensation in LVLMs.
More closely related, Fourier-VLM~\cite{wang2025fourier} applies 2D DCT to visual features and truncates high-frequency coefficients.
Its compression is spatially uniform and query-agnostic, does not address temporal budget allocation for long videos, and requires additional fine-tuning.

In contrast, \ours adopts frequency-aware decomposition as the organizing principle of compression.
Rather than pruning tokens directly or applying uniform frequency truncation, it separates spatiotemporal signals into components with different functional roles and performs query-conditioned condensation in the decomposed space.
This formulation treats temporal redundancy and spatial detail preservation as two coupled but distinct objectives. 
Unlike Fourier-VLM~\cite{wang2025fourier}, \ours uses localized wavelet decomposition, explicitly models both temporal and spatial compression, and requires no task-specific training.

\section{Method}

\subsection{Problem Formulation}
Consider a video comprising \(T\) frames \(\{I_t\}_{t=1}^T\) and a text query \(Q=\{q_j\}_{j=1}^{L}\).
A LVLM first encodes each frame with a vision encoder \(E_v\):
\begin{equation}
X_t = E_v(I_t) \in \mathbb{R}^{C\times H\times W},
\end{equation}
and we denote the full set of frame features as \(X=\{X_t\}_{t=1}^T\).
Spatially flattening each feature map yields \(N=H\times\!W\) patch tokens \(\{x_{t,n}\}_{n=1}^{N}\), projected to LLM hidden dimension \(d\):
\begin{equation}
z_{t,n} = P(x_{t,n}) \in \mathbb{R}^{d}, \quad n=1,\dots,N.
\end{equation}
Concatenating all visual tokens across frames gives
\begin{equation}
Z = [z_{1,1},\dots,z_{T,N}] \in \mathbb{R}^{N_v\times d}, \quad N_v = T \cdot N.
\end{equation}
During prefilling, the model processes the joint sequence \([Z;Q]\), and the self-attention cost scales as \(\mathcal{O}\!\big((N_v+L)^2\big)\).
Since \(N_v \gg L\) for long videos, the central goal is to reduce \(N_v\) while preserving the visual evidence most relevant to the query. We formalize this as \emph{spatial-temporal token condensation}, i.e., a query-conditioned mapping:
\begin{equation}
\tilde{Z} = \mathcal{C}(X, Q) \in \mathbb{R}^{B\times d}, \quad B \ll N_v.
\end{equation}
The total budget \(B\) is distributed across frames via allocations \(\{n_t\}_{t=1}^{T}\) subject to:
\begin{equation}
\sum_{t=1}^{T} n_t = B, \quad n_t \in \mathbb{Z}_{\ge 0},
\end{equation}
thereby allowing variable token counts across frame.
To quantify cross-modal relevance, we ues a lightweight pretrained cross-modal model to compute two complementary signals: (i) frame-level relevance scores and (ii) token-level spatial saliency maps.
This module uses only its cross-modal encoder and regression head without any decoder-stage computation, and can alternatively be instantiated using base LVLM features.
Given a fixed pretrained LVLM $F$, we seek an operator $\mathcal{C}$ that requires no task-specific optimization and satisfies $F(\widetilde{Z},Q)\approx F(Z,Q)$ while substantially reducing the prefilling cost.

\label{sec:method}
\begin{figure*}[!t]
    \centering
    \includegraphics[width=\textwidth]{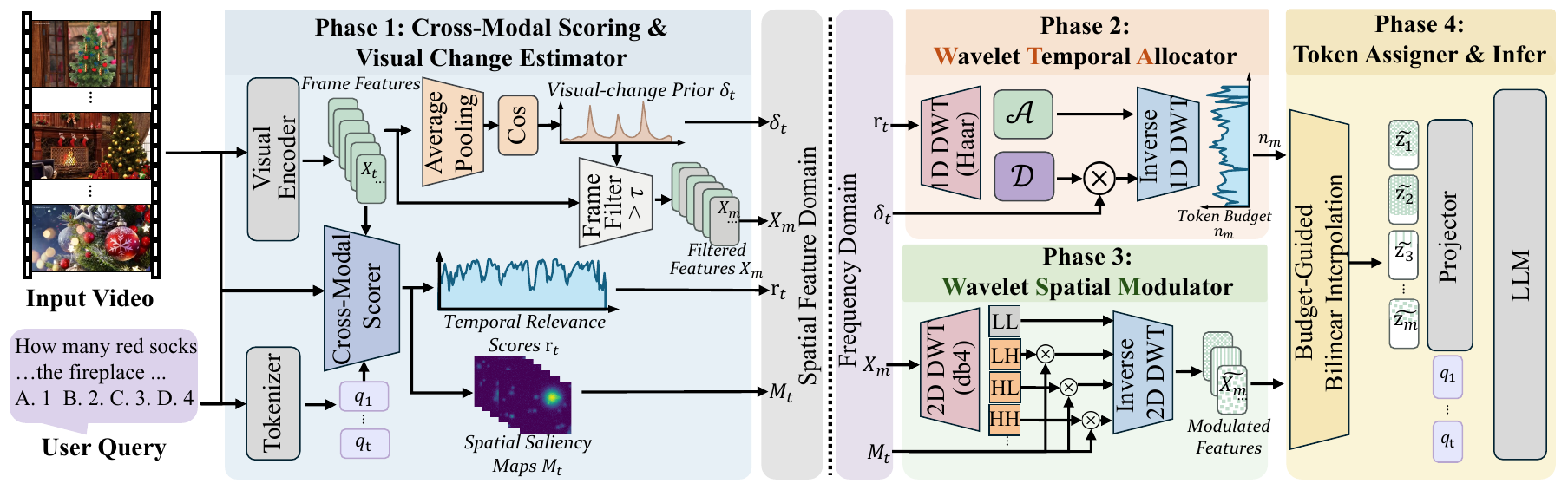}
    \caption{\textbf{Overview of \ours.} 
    The pipeline computes cross-modal relevance, rectifies temporal scores via WTA (Wavelet Temporal Allocator), preserves spatial saliency via WSM (Wavelet Spatial Modulator), and performs budget-guided compression.}
    \label{fig:pipeline}
\end{figure*}

\subsection{\ours}
We introduce \ours, a training-free pipeline for spatial-temporal token condensation in LVLMs (Fig.~\ref{fig:pipeline}). 
The pipeline consists of four sequential modules: the Cross-Modal Scorer, the Visual Change Estimator (VCE), the Wavelet Processor, and the Token Assigner.
Given video frames \(\{I_t\}_{t=1}^{T}\) and a text query \(Q\), the Cross-Modal Scorer first produces frame-level relevance scores \(r_t\) and token-level spatial saliency maps \(M_t\). 
The VCE estimates inter-frame visual changes, yielding a temporal gate \(\gamma_t\), and filters highly redundant frames to form the retained index set \(\mathcal{M}\). 
The Wavelet Processor then operates on the retained sequence: WTA rectifies the relevance sequence \(\{r_m\}_{m\in\mathcal{M}}\) for dynamic frame-level budget allocation, while WSM modulates high-frequency spatial subbands of \(\{X_m\}_{m\in\mathcal{M}}\) using saliency maps \(\{M_m\}_{m\in\mathcal{M}}\). 
Finally, the Token Assigner compresses each retained feature map according to its allocated budget and concatenates the projected tokens into the condensed sequence:
\begin{equation}
\widetilde Z = \mathcal{C}_{\mathrm{WaveZip}}(X,Q)\in\mathbb{R}^{B\times d}.
\end{equation}

\subsubsection{Cross-Modal Scorer}
We employ a lightweight, frozen cross-modal encoder $\Phi$ with visual branch $\phi_v(\cdot)$ and textual branch $\phi_q(\cdot)$ to compute query-conditioned relevance without additional parameter updates.
The relevance scorer can be instantiated either through embedding similarity or through a pretrained image-text matching model.
For frame $I_t$ and query $Q$, we extract patch-level visual embeddings $u_{t,n} = \phi_v(I_t)_n \in \mathbb{R}^{D}$ for $n=1,\dots,N$ and the global query embedding $u_q = \phi_q(Q) \in \mathbb{R}^{D}$.
\noindent
\textbf{Temporal Relevance.}
A generic scoring function $\mathcal{F}_{\text{time}}$ quantifies the global semantic alignment between frame $I_t$ and query $Q$, yielding a scalar relevance score $r_t \in \mathbb{R}$.
In the \emph{similarity-based} branch, $\mathcal{F}_{\text{time}}$ aggregates visual tokens via average pooling and computes cosine similarity with $u_q$.
In the \emph{matcher-based} branch, $r_t$ is directly given by the ITM head output of the cross-modal model.
\noindent
\textbf{Spatial Saliency.}
A token-level scoring function $\mathcal{F}_{\text{space}}$ produces a patch-wise relevance score $a_{t,n} \in \mathbb{R}$ for each visual token $u_{t,n}$, realized via token-wise cosine similarity with $u_q$ or by extracting cross-attention weights within the ITM model.
These scores are reshaped to the 2D spatial grid and processed by a rank-preserving normalization $\mathcal{N}$ to yield the spatial saliency map:
\begin{equation}
    M_t = \mathcal{N}\Big(\mathcal{R}_{1\text{D}\to 2\text{D}}\big(\{a_{t,n}\}_{n=1}^{N}\big)\Big) \in [0, 1]^{H\times W}.
\end{equation}
Together, $r_t$ and $M_t$ provide cross-modal guidance for the subsequent wavelet-based condensation: 
$r_t$ drives WTA, while $M_t$ guides spatial detail reweighting within WSM.

\subsubsection{Visual Change Estimator (VCE)}
 
To distinguish structural scene transitions from spurious inter-frame jitter, the Visual Change Estimator (VCE) establishes a visual-change prior.
The inter-frame temporal change is quantified as the cosine distance between the spatially average-pooled features of adjacent frames:
\begin{equation}
    \delta_t = 1 - \frac{\bar{v}_t \cdot \bar{v}_{t-1}}{\|\bar{v}_t\|_2 \|\bar{v}_{t-1}\|_2} \in [0, 2], \quad t=2,\ldots,T, \quad \delta_1 = 0,
\end{equation}
where $\bar{v}_t$ denotes the spatially average-pooled feature of frame $X_t$.
Applying min-max normalization over the temporal dimension yields the gating factor $\gamma_t = \mathcal{N}_{\text{min-max}}(\delta_t) \in [0, 1]$,
where $\gamma_t \to 1$ signifies an abrupt scene cut and $\gamma_t \to 0$ denotes a static segment. 
This factor serves as a critical condition for the subsequent temporal allocation.

In addition, a Frame Filter is incorporated within the VCE to remove highly redundant adjacent frames. 
Specifically, if the visual change $\delta_t$ falls below a predefined threshold $\tau$, the corresponding frame is considered redundant and is subsequently removed.
We denote the retained frame index set as
\(\mathcal{M}=\{t\mid \delta_t\ge \tau\}\).
The retained-frame signals are written as
\(\{X_m,r_m,M_m,\gamma_m\}_{m\in\mathcal{M}}\), where \(\mathcal{M}\) preserves the original temporal order. 
All subsequent wavelet processing and token assignment are performed on this retained sequence.

\subsubsection{Wavelet Processor}
The Wavelet Processor rectifies the temporal relevance sequence and reweights spatial detail components through two parallel branches: the Wavelet Temporal Allocator (WTA) and the Wavelet Spatial Modulator (WSM).
\noindent\textbf{Wavelet Temporal Allocator (WTA).}
The WTA stabilizes the temporal relevance sequence \(R_{\mathcal{M}}=[r_m]_{m\in\mathcal{M}}\) by separating its global relevance trend from local fluctuations.
We apply a single-level 1D DWT with the Haar basis:
\begin{equation}
(A,D)=\operatorname{DWT}_{\mathrm{Haar}}(R_{\mathcal{M}}),
\end{equation}
where \(A\) contains low-frequency approximation coefficients and \(D\) contains high-frequency detail coefficients.
Since the length of \(D\) differs from the original retained sequence, the visual-change gate \(\Gamma_{\mathcal{M}}=[\gamma_m]_{m\in\mathcal{M}}\) is first downsampled to the detail-coefficient resolution:
\begin{equation}
\bar{\Gamma}_{\mathcal{M}}=\mathcal{P}_{\downarrow}(\Gamma_{\mathcal{M}}),
\qquad
\widetilde{D}=\bar{\Gamma}_{\mathcal{M}}\odot D .
\end{equation}
This operation attenuates high-frequency relevance fluctuations in visually static segments while preserving detail coefficients around frames with larger visual changes.
The rectified signal is reconstructed by inverse DWT:
\begin{equation}
\widetilde{R}_{\mathcal{M}}
=
[\widetilde r_m]_{m\in\mathcal{M}}
=
\operatorname{IDWT}_{1\mathrm{D}}(A,\widetilde{D}).
\end{equation}

\noindent\textbf{Wavelet Spatial Modulator (WSM).}
The WSM reweights query-relevant spatial details by modulating the high-frequency wavelet subbands while keeping the low-frequency approximation unchanged.
For each retained feature map \(X_m\in\mathbb{R}^{C\times H\times W}\), we apply a single-level 2D DWT with the db4 basis:
\begin{equation}
(X_m^{LL},X_m^{LH},X_m^{HL},X_m^{HH})
=
\operatorname{DWT}_{2\mathrm{D}}(X_m),
\end{equation}
where \(LL\) captures low-frequency structure and \(\{LH,HL,HH\}\) encode high-frequency details.
We construct a saliency-guided modulation gate
\begin{equation}
G_m=\lambda\cdot \mathcal{P}_{\downarrow}(M_m),
\end{equation}
where \(\mathcal{P}_{\downarrow}\) average-pools the saliency map to the wavelet subband resolution and \(\lambda\) controls the modulation strength.
The gate is applied only to high-frequency subbands:
\begin{equation}
\widetilde X_m^{K}=X_m^{K}\odot G_m,
\qquad K\in\{LH,HL,HH\}.
\end{equation}
The modulated feature map is reconstructed by inverse 2D DWT:
\begin{equation}
\widetilde X_m
=
\operatorname{IDWT}_{2\mathrm{D}}
(X_m^{LL},\widetilde X_m^{LH},\widetilde X_m^{HL},\widetilde X_m^{HH}).
\end{equation}

\begin{table*}[!t]
\vspace{-2mm}
\centering
\small
\setlength{\tabcolsep}{10pt}
\renewcommand{\arraystretch}{0.95}
\begin{tabular}{l c ccc cc}
\toprule
\multirow{2}{*}{Method} &
\multirow{2}{*}{\makecell{Retained\\Ratio $\rho$}} &
\multicolumn{3}{c}{Benchmarks (\%) $\uparrow$} &
\multicolumn{2}{c}{Average $\uparrow$} \\
\cmidrule(lr){3-5} \cmidrule(lr){6-7}
& & EgoSchema & LongVideoBench & VideoMME & Score & \% \\
\midrule

\rowcolor{gray!20}
LLaVA-OV-7B~\cite{li2024llava}
& 100\%
& 62.1 & 56.4 & 58.6 & 59.0 & 100.0 \\

\midrule
FastV~\cite{chen2024image}
& 25\%
& 60.4 & 56.7 & 56.1 & 57.7 & 97.8 \\

VisionZip~\cite{yang2025visionzip}
& 25\%
& 63.0 & 56.5 & 58.2 & 59.2 & 100.3 \\

FastVID~\cite{shen2025fastvid}
& 25\%
& 61.2 & 55.9 & 57.9 & 58.3 & 98.8 \\

DyCoke~\cite{tao2025dycoke}
& 25\%
& 64.0 & 55.7 & 59.5 & 59.7 & 101.2 \\

PruMerge~\cite{shang2025llava}
& 25\%
& \textbf{64.6} & 56.1 & 57.4 & 59.4 & 100.6 \\

VFlowOpt~\cite{yang2025vflowopt}
& 25\%
& 64.4 & 56.3 & 57.7 & 59.5 & 100.8 \\

\rowcolor{cyan!10}
\textbf{WaveZip}
& 25\%
& 63.6 & \textbf{57.5} & \textbf{61.1} & \textbf{60.7} & \textbf{102.8} \\

\midrule
FastV~\cite{chen2024image}
& 20\%
& 60.6 & 55.9 & 56.9 & 57.8 & 98.0 \\

VisionZip~\cite{yang2025visionzip}
& 20\%
& 62.0 & 55.2 & 57.9 & 58.4 & 98.9 \\

FastVID~\cite{shen2025fastvid}
& 20\%
& 61.2 & 55.9 & 58.0 & 58.4 & 98.9 \\

DyCoke~\cite{tao2025dycoke}
& 20\%
& \textbf{65.0} & 54.8 & 58.5 & 59.4 & 100.7 \\

PruMerge~\cite{shang2025llava}
& 20\%
& 64.0 & 55.5 & 56.5 & 58.7 & 99.4 \\

VFlowOpt~\cite{yang2025vflowopt}
& 20\%
& 64.0 & 56.1 & 58.2 & 59.4 & 100.7 \\

\rowcolor{cyan!10}
\textbf{WaveZip}
& 20\%
& 63.0 & \textbf{58.1} & \textbf{60.8} & \textbf{60.6} & \textbf{102.7} \\

\midrule
FastV~\cite{chen2024image}
& 15\%
& 59.8 & 54.8 & 54.6 & 56.4 & 95.5 \\

VisionZip~\cite{yang2025visionzip}
& 15\%
& 62.8 & 54.4 & 56.1 & 57.8 & 97.8 \\

FastVID~\cite{shen2025fastvid}
& 15\%
& 58.8 & \textbf{56.7} & 57.7 & 57.7 & 97.8 \\

DyCoke~\cite{tao2025dycoke}
& 15\%
& \textbf{63.8} & 54.7 & 58.0 & 58.8 & 99.7 \\

PruMerge~\cite{shang2025llava}
& 15\%
& 63.4 & 54.5 & 55.8 & 57.9 & 98.2 \\

VFlowOpt~\cite{yang2025vflowopt}
& 15\%
& \textbf{63.8} & 54.5 & 58.2 & 58.8 & 99.7 \\

\rowcolor{cyan!10}
\textbf{WaveZip}
& 15\%
& 63.6 & 55.5 & \textbf{59.5} & \textbf{59.5} & \textbf{100.8} \\

\midrule
FastV~\cite{chen2024image}
& 10\%
& 59.0 & 52.4 & 52.7 & 54.7 & 92.6 \\

VisionZip~\cite{yang2025visionzip}
& 10\%
& 61.6 & 49.3 & 53.4 & 54.8 & 92.8 \\

FastVID~\cite{shen2025fastvid}
& 10\%
& 58.8 & \textbf{55.7} & 57.2 & 57.2 & 96.9 \\

DyCoke~\cite{tao2025dycoke}
& 10\%
& \textbf{63.0} & 52.9 & 57.1 & 57.7 & 97.7 \\

PruMerge~\cite{shang2025llava}
& 10\%
& 61.0 & 55.1 & 54.9 & 57.0 & 96.6 \\

VFlowOpt~\cite{yang2025vflowopt}
& 10\%
& 62.0 & 55.1 & 56.7 & 57.9 & 98.2 \\

\rowcolor{cyan!10}
\textbf{WaveZip}
& 10\%
& 62.3 & 54.7 & \textbf{59.6} & \textbf{58.8} & \textbf{99.6} \\

\bottomrule
\end{tabular}
\caption{\textbf{Main comparison on LLaVA-OneVision-7B.}
EgoSchema results are evaluated on the 500-sample subset. 
\textbf{Best} results under the same retained ratio are highlighted in bold. 
All scores are accuracy (\%).}
\label{tab:ov_benchmark}
\vspace{-2mm}
\end{table*}

\subsubsection{Token Assigner}
The Token Assigner allocates the global token budget according to the rectified retained-frame relevance.
For each retained frame \(m\in\mathcal{M}\), the token budget is assigned as
\begin{equation}
n_m =
\operatorname{Round}
\left(
B\cdot
\frac{\max(\widetilde r_m,0)}
{\sum_{k\in\mathcal{M}}\max(\widetilde r_k,0)}
\right),
\qquad m\in\mathcal{M}.
\end{equation}
For frames with \(n_m>0\), the modulated feature map \(\widetilde X_m\in\mathbb{R}^{C\times H\times W}\) is first compressed into \(n_m\) feature tokens through budget-guided bilinear interpolation:
\begin{equation}
\widehat X_m=\Psi(\widetilde X_m,n_m)\in\mathbb{R}^{n_m\times C}.
\end{equation}
The interpolated features are then mapped into the LLM hidden space by the pretrained projector \(P\):
\begin{equation}
\widetilde z_m=P(\widehat X_m)\in\mathbb{R}^{n_m\times d}.
\end{equation}
Finally, the retained-frame tokens are concatenated in temporal order:
\begin{equation}
\widetilde Z
=
\operatorname{Concat}_{m\in\mathcal{M}}(\widetilde z_m)
\in\mathbb{R}^{B\times d},
\end{equation}
which is directly fed into the LVLM for efficient prefilling.

\section{Experiment}
\label{sec:experiment}

\begin{table*}[!t]
\centering
\small
\setlength{\tabcolsep}{5.5pt}
\renewcommand{\arraystretch}{0.92}
\begin{tabular}{c|l|c|ccc|cc}
\toprule
\multirow{2}{*}{Backbone} &
\multirow{2}{*}{Method} &
\multirow{2}{*}{\makecell{Retained\\Ratio $\rho$}} &
\multicolumn{3}{c|}{Benchmarks (\%) $\uparrow$} &
\multicolumn{2}{c}{Average $\uparrow$} \\
\cmidrule(lr){4-6} \cmidrule(lr){7-8}
& & & EgoSchema & LongVideoBench & VideoMME & Score & \% \\
\midrule

\multirow{9}{*}{\makecell{LLaVA-\\Video-7B\\~\cite{zhang2024video}}}
& \cellcolor{gray!20}{Vanilla}
& \cellcolor{gray!20}{100\%}
& \cellcolor{gray!20}{57.2}
& \cellcolor{gray!20}{58.9}
& \cellcolor{gray!20}{64.3}
& \cellcolor{gray!20}{60.1}
& \cellcolor{gray!20}{100.0} \\
\cmidrule{2-8}

& FastV~\cite{chen2024image}
& 20\%
& 54.8 & 56.0 & 59.2 & 56.6 & 94.2 \\

& VisionZip~\cite{yang2025visionzip}
& 20\%
& 59.0 & 58.0 & 61.7 & 59.6 & 99.1 \\

& FastVID~\cite{shen2025fastvid}
& 20\%
& 57.0 & 57.1 & 62.6 & 58.9 & 98.0 \\

& \cellcolor{cyan!10}{\textbf{WaveZip}}
& \cellcolor{cyan!10}{20\%}
& \cellcolor{cyan!10}{\textbf{61.4}}
& \cellcolor{cyan!10}{\textbf{58.2}}
& \cellcolor{cyan!10}{\textbf{62.7}}
& \cellcolor{cyan!10}{\textbf{60.7}}
& \cellcolor{cyan!10}{\textbf{101.0}} \\
\cmidrule{2-8}

& FastV~\cite{chen2024image}
& 10\%
& 50.6 & 53.6 & 55.8 & 53.3 & 88.7 \\

& VisionZip~\cite{yang2025visionzip}
& 10\%
& 54.4 & 54.5 & 59.5 & 56.1 & 93.4 \\

& FastVID~\cite{shen2025fastvid}
& 10\%
& 54.8 & 56.3 & 59.6 & 56.9 & 94.7 \\

& \cellcolor{cyan!10}{\textbf{WaveZip}}
& \cellcolor{cyan!10}{10\%}
& \cellcolor{cyan!10}{\textbf{59.0}}
& \cellcolor{cyan!10}{\textbf{57.0}}
& \cellcolor{cyan!10}{\textbf{61.1}}
& \cellcolor{cyan!10}{\textbf{59.0}}
& \cellcolor{cyan!10}{\textbf{98.1}} \\

\midrule

\multirow{9}{*}{\makecell{Qwen2.5-VL\\~\cite{bai2025qwen2}}}
& \cellcolor{gray!20}{Vanilla}
& \cellcolor{gray!20}{100\%}
& \cellcolor{gray!20}{61.6}
& \cellcolor{gray!20}{55.3}
& \cellcolor{gray!20}{61.3}
& \cellcolor{gray!20}{59.4}
& \cellcolor{gray!20}{100.0} \\
\cmidrule{2-8}

& VisionZip~\cite{yang2025visionzip}
& 20\%
& \textbf{62.0} & 51.8 & 56.5 & 56.7 & 95.4 \\

& FastVID~\cite{shen2025fastvid}
& 20\%
& 60.5 & 51.6 & 57.5 & 56.5 & 95.2 \\

& FlashVID~\cite{fan2026flashvid}
& 20\%
& 59.6 & OOM & OOM & -- & -- \\

& \cellcolor{cyan!10}{\textbf{WaveZip}}
& \cellcolor{cyan!10}{20\%}
& \cellcolor{cyan!10}{59.6}
& \cellcolor{cyan!10}{\textbf{54.2}}
& \cellcolor{cyan!10}{\textbf{58.3}}
& \cellcolor{cyan!10}{\textbf{57.4}}
& \cellcolor{cyan!10}{\textbf{96.6}} \\
\cmidrule{2-8}

& VisionZip~\cite{yang2025visionzip}
& 10\%
& \textbf{61.0} & 49.1 & 54.9 & 55.0 & 92.5 \\

& FastVID~\cite{shen2025fastvid}
& 10\%
& 58.4 & 50.2 & 55.3 & 54.6 & 91.9 \\

& FlashVID~\cite{fan2026flashvid}
& 10\%
& 58.0 & OOM & OOM & -- & -- \\

& \cellcolor{cyan!10}{\textbf{WaveZip}}
& \cellcolor{cyan!10}{10\%}
& \cellcolor{cyan!10}{57.4}
& \cellcolor{cyan!10}{\textbf{53.6}}
& \cellcolor{cyan!10}{\textbf{56.4}}
& \cellcolor{cyan!10}{\textbf{55.8}}
& \cellcolor{cyan!10}{\textbf{93.9}} \\

\bottomrule
\end{tabular}
\caption{\textbf{Backbone generalization of WaveZip.}
We further evaluate WaveZip on LLaVA-Video-7B and Qwen2.5-VL. EgoSchema results are evaluated on the 500-sample subset. 
\textbf{Best} results under the same backbone and retained ratio are highlighted in bold. 
``OOM'' indicates out-of-memory under the same hardware and data setting.}
\label{tab:backbone_generalization}
\end{table*}

\subsection{Experimental Setup}
\label{sec:exp_settings}
%
We conduct evaluations across three benchmarks: EgoSchema~\cite{mangalam2023egoschema}, LongVideoBench~\cite{wu2024longvideobench}, and VideoMME~\cite{fu2025video}.
These benchmarks are widely adopted in video understanding research and span diverse video lengths and task complexities, providing a comprehensive testbed for evaluating compression methods under various scenarios.
For the main comparison, we use LLaVA-OneVision-7B~\cite{li2024llava}, where the most complete set of recent token compression baselines is available. 
To examine backbone generalization, we further integrate WaveZip into LLaVA-Video-7B~\cite{zhang2024video} and Qwen2.5-VL~\cite{bai2025qwen2}. 
For each backbone, we evaluate multiple retained-token ratios, with the total token budget set to \(B=\rho N_v\).

All methods use 64 uniformly sampled frames unless otherwise specified.
We use BLIP-ITM-LARGE-COCO~\cite{li2022blip} as the frozen cross-modal scorer, Haar for WTA, db4 for WSM, and $\lambda=1.5$ by default.
All experiments are conducted on NVIDIA A800 GPUs.

%
%
%
%
\subsection{Main Results}
\textbf{Main comparison on LLaVA-OneVision.}
Table~\ref{tab:ov_benchmark} reports the main comparison on LLaVA-OneVision-7B across four retained-token ratios. 
WaveZip achieves the best average accuracy at every retained-token ratio without task-specific training, outperforming both classical pruning or merging baselines and recent video-oriented methods such as DyCoke, PruMerge, and VFlowOpt.
This advantage remains stable as the token budget decreases: even under the aggressive $\rho=0.1$ setting, WaveZip retains 99.6\% of the full-token performance.
Beyond the averaged results, WaveZip achieves the highest VideoMME accuracy at all four retained-token ratios.
These results demonstrate a robust accuracy--compression trade-off across different budgets.

\par
\noindent
\textbf{Generalization across LVLM backbones.}
Table~\ref{tab:backbone_generalization} further evaluates WaveZip on LLaVA-Video-7B and Qwen2.5-VL. 
On LLaVA-Video-7B, WaveZip achieves the
highest average score at both retained ratios.
On LLaVA-Video-7B, WaveZip achieves the best average performance at both \(\rho=0.2\%\) and \(\rho=0.1\%\), preserving 101.0\% and 98.1\% of the full-input performance, respectively. 
On Qwen2.5-VL, WaveZip also obtains the highest average score under both retained ratios, with consistent gains on LongVideoBench and VideoMME. 
These results support the plug-and-play nature of WaveZip: the same frequency-domain condensation principle transfers across different LVLMs without task-specific training.

\subsection{Ablation Study}
\begin{figure}[!t]
    \centering
    
    \includegraphics[width=1.0\linewidth]{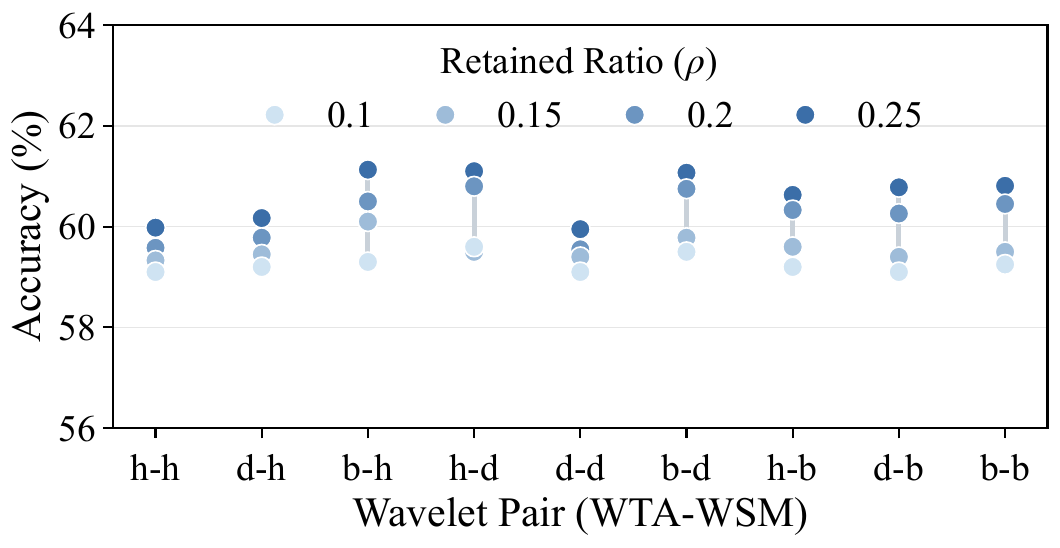}
    \caption{\textbf{Ablation of wavelet bases in WTA and WSM on VideoMME with LLaVA-OV.} 
    Each group on the x-axis denotes a basis pair assigned to WTA and WSM, respectively (h: Haar, d: db4, b: bior4.4).
    Each dot represents the accuracy at a specific retained ratio ($\rho \in \{0.10, 0.15, 0.20, 0.25\}$), and the vertical segment spans the min--max range across all four ratios.}
    \label{fig:ablation_wavelet}
\end{figure}
%

\noindent
\textbf{Wavelet Basis.}
We evaluate three wavelet bases, including Haar, db4, and bior4.4, while keeping the decomposition level fixed to one.
As shown in Fig.~\ref{fig:ablation_wavelet}, we observe no consistent winner across settings: the top-ranked basis pair varies with the retained ratios, and the relative ordering among pairs remains unstable. 
For any fixed retained ratio, accuracy values are tightly clustered, with an average range of approximately 0.8\%, indicating only marginal sensitivity to the specific wavelet family.
This near-uniform pattern suggests that no single wavelet family offers a systematic advantage, and the performance gains of WaveZip arise primarily from the multi-band decomposition mechanism itself rather than from any particular basis selection.
\begin{figure}[!t]
    \centering
    \includegraphics[width=1.0\linewidth]{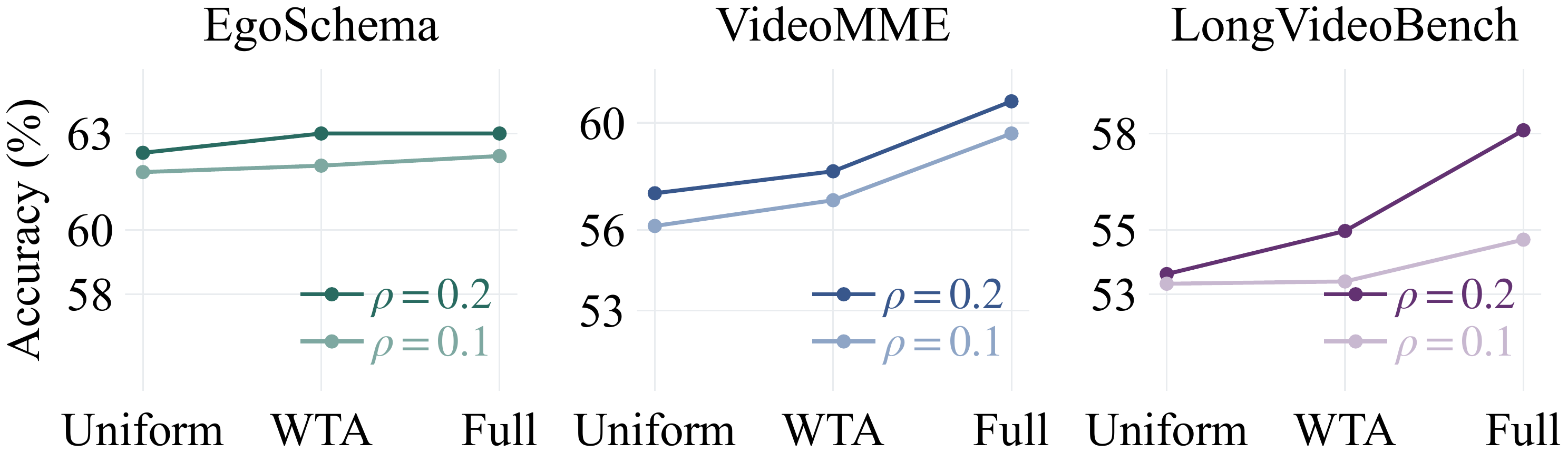}
    \caption{\textbf{Ablation of core modules in WaveZip.}
    \textit{Uniform} denotes the uniform compression baseline, \textit{WTA} adds only the Wavelet Temporal Allocator, and \textit{Full} further adds the Wavelet Spatial Modulator. 
    Results are reported at two retained ratios, \(\rho=0.1\) and \(\rho=0.2\).}
    \label{fig:module_ablation}
\end{figure}
\begin{figure}[!t] 
    \centering
    \includegraphics[width=0.95\linewidth]{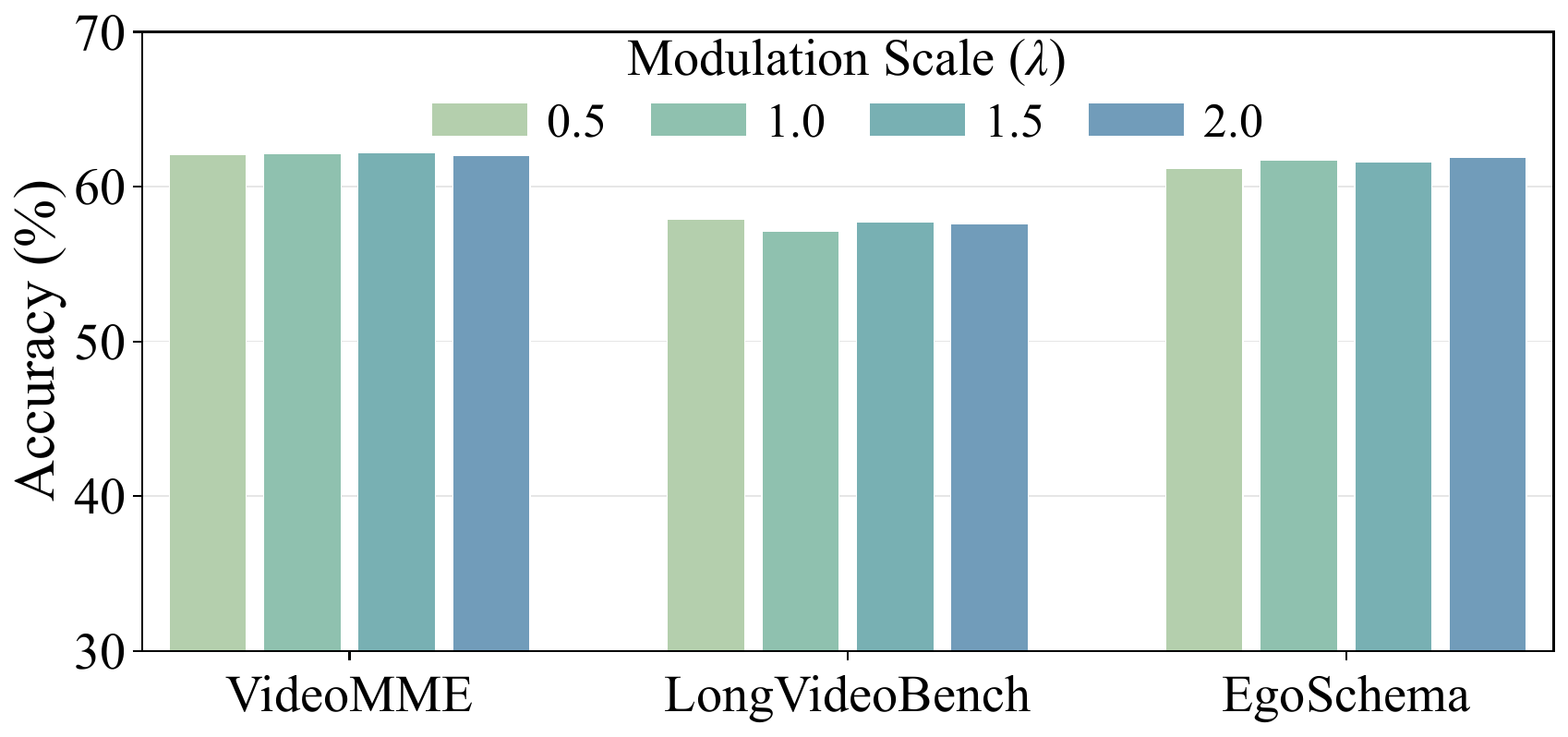}
    \caption{\textbf{Ablation of the WSM modulation scale \(\lambda\) on LLaVA-OV.}
    Bars report accuracy for \(\lambda\in\{0.5,1.0,1.5,2.0\}\), showing that performance is relatively stable with only minor variations.}
    \label{fig:ablation_gate_scale}
\end{figure}

\noindent
\textbf{Effectiveness of Core Modules.}
To assess the contribution of each component, we compare three configurations: \textit{Uniform}, \textit{WTA}, and \textit{Full}. 
\textit{Uniform} denotes a uniform compression baseline, \textit{WTA} adds only the Wavelet Temporal Allocator, and \textit{Full} further incorporates the Wavelet Spatial Modulator. 
As shown in Fig.~\ref{fig:module_ablation}, WTA generally improves over uniform compression by introducing frequency-rectified temporal budget allocation. 
Adding WSM provides clear further gains on VideoMME and LongVideoBench, while the performance on EgoSchema changes only marginally.
The full model achieves the strongest overall performance, indicating that temporal allocation and spatial modulation provide complementary benefits across the benchmark suite. 
\begin{table}[!t]
\centering
\footnotesize
\setlength{\tabcolsep}{7pt}
\begin{tabular}{lcccc}
\toprule
Method & 16 Frames & 32 Frames & 96 Frames & 128 Frames \\
\midrule
FastV     & 48.8 & 53.1 & OOM  & OOM \\
VisionZip & 47.1 & 47.6 & 48.6 & OOM \\
FastVID   & 54.5 & 57.3 & 59.9 & 59.7 \\
WaveZip   & \textbf{54.8} & \textbf{58.7} & \textbf{61.9} & \textbf{62.1} \\
\bottomrule
\end{tabular}
\caption{\textbf{Ablation with different frame budgets on VideoMME using LLaVA-Video.}
All compressed methods use the same retained-token ratio of \(\rho=0.2\) and are evaluated under the same hardware setting. 
``OOM'' denotes out-of-memory.
}
\label{tab:frame_budget_ablation}
\end{table}

\noindent
\textbf{Different Frame Budgets.}
The main experiments use 64 uniformly sampled frames. We further vary the maximum number of sampled frames on VideoMME with LLaVA-Video to evaluate robustness to different input lengths. 
All compressed methods use the same retained-token ratio of \(\rho=0.2\) under each frame setting.
As shown in Table~\ref{tab:frame_budget_ablation}, WaveZip achieves the highest accuracy from 16 to 128 frames. Its accuracy increases from 54.8\% with 16 frames to 62.1\% with 128 frames, while FastV and VisionZip encounter out-of-memory errors at larger frame budgets.
WaveZip also consistently outperforms FastVID, with the margin
increasing from 0.3 points at 16 frames to 1.4, 2.0, and 2.4 points at 32, 96, and 128 frames, respectively. 
These results show that WaveZip remains effective as the sampled video sequence becomes longer and is not tailored to the default 64-frame setting.

%

\begin{table}[!t]
\centering
\small
\setlength{\tabcolsep}{4pt}
\begin{tabular}{lccc}
\toprule
Method 
& \textbf{VideoMME} 
& \textbf{\shortstack{LongVideoBench}} 
& \textbf{EgoSchema} \\
\midrule
Similarity & 62.0 & 57.5 & 62.5 \\
BLIP       & 62.7 & 58.2 & 61.4 \\
BLIP2      & 62.3 & 57.7 & 61.0 \\
\bottomrule
\end{tabular}
\caption{\textbf{Ablation of cross-modal module on LLaVA-Video-7B.}
Similarity denotes using the base LVLM embeddings to compute \(r_t\) and \(M_t\).
All scores are accuracy (\%).}
\label{tab:ablation_xmod}
\end{table}

\noindent
\textbf{Spatial Modulation Scale.}
We ablate the WSM modulation scale \(\lambda\), which controls the strength of saliency-guided high-frequency modulation.
Fig.~\ref{fig:ablation_gate_scale} shows the performance variations across different \(\lambda\) values are small on all benchmarks that WaveZip is not sensitive to the exact modulation scale. We use $\lambda=1.5$ as default across all experiments.

%
\noindent
\textbf{Cross-Modal Relevance Modeling.}
We compare three strategies for computing the cross-modal relevance signals $r_t$ and $M_t$: direct similarity using the base LVLM embeddings, BLIP, and BLIP2. 
As shown in Table~\ref{tab:ablation_xmod}, BLIP performs slightly better on VideoMME and
LongVideoBench, whereas direct similarity achieves the highest accuracy on EgoSchema. 
The overall differences among the three scorers remain small, and no single scorer dominates all benchmarks. 
These results indicate that WaveZip is not tightly coupled to a particular relevance estimator: its temporal rectification and spatial frequency modulation remain effective with either base-LVLM similarity or external image-text matching models.

\begin{figure}[!t]
    \centering
    
    \includegraphics[width=\columnwidth]{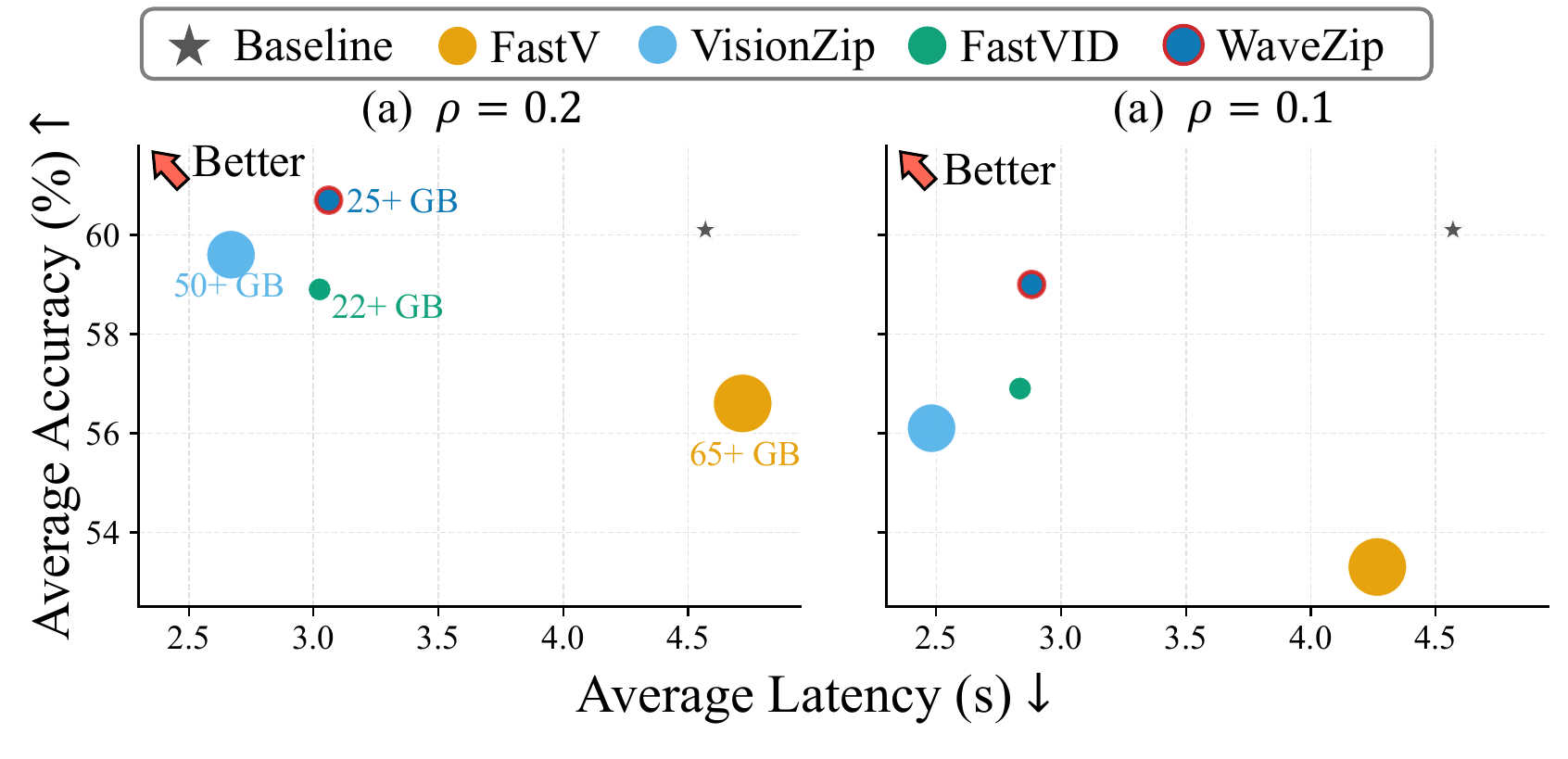}
    \caption{\textbf{End-to-end accuracy--efficiency trade-off across three benchmarks with LLaVA-Video-7B.}
    Average latency and peak GPU memory are measured separately on a fixed 600-sample subset constructed by sampling 200 examples from each benchmark. 
    Results are reported at \(\rho=0.1\) and \(\rho=0.2\), with marker size indicating peak GPU memory. 
    Latency covers input preprocessing, cross-modal scoring, token compression, LVLM prefilling, and response generation.}
    \label{fig:efficiency_tradeoff}
\end{figure}

\subsection{Efficiency Analysis}
We evaluate the end-to-end inference efficiency of different token compression methods with LLaVA-Video-7B. 
The accuracy coordinates in Fig.~\ref{fig:efficiency_tradeoff} use the benchmark-level results reported in Table~\ref{tab:backbone_generalization}. 
Latency and peak GPU memory are measured separately on a fixed 600-sample subset constructed by sampling 200 examples from each of the three benchmarks.
Each sample is processed independently and serially with a batch size of one.
Latency is measured from input preprocessing to final response generation.
The cross-modal scorer and the LVLM are executed sequentially without parallelization or cross-query feature caching.
As shown in Fig.~\ref{fig:efficiency_tradeoff}, WaveZip achieves a favorable trade-off among average accuracy, latency, and peak GPU memory. 
At $\rho=0.1$, WaveZip obtains an average score of 59.0 while achieving a $1.59\times$ end-to-end speedup.
It has comparable latency to FastVID and improves the average score by 2.1 percentage points. Compared with VisionZip, WaveZip improves the average score by 2.9 points while reducing peak GPU memory from 48.8 GB to 23.9 GB.
At \(\rho=0.2\), WaveZip achieves an average score of 60.7, exceeding the full-token result by 0.6 points and the strongest compressed baseline by 1.1 points, while providing a \(1.48\times\) speedup.
These results show that WaveZip substantially reduces end-to-end inference cost while maintaining competitive average accuracy across the evaluated long-video benchmarks.

\begin{figure}[!t]
    \centering
    
    \includegraphics[width=\columnwidth]{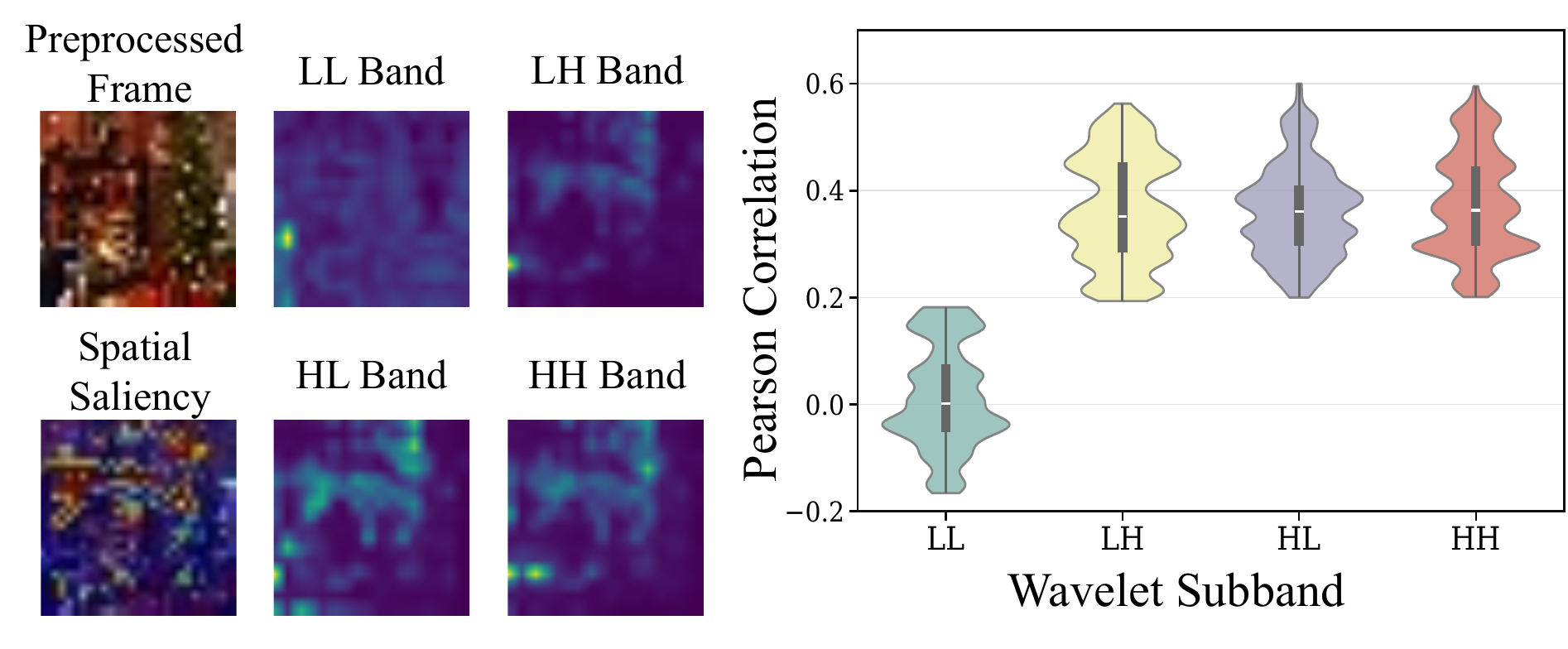}
    \caption{
    \textbf{Frequency-band saliency analysis.}
    Wavelet-band energy maps and Pearson correlation statistics show that query-conditioned saliency aligns more strongly with high-frequency detail bands than with the low-frequency LL band, motivating saliency-guided high-frequency modulation in WSM.
    }
    \label{fig:band_saliency_analysis}
\end{figure}

\subsection{Frequency-Band Saliency Analysis}
We analyze whether query-conditioned saliency is more aligned with wavelet detail components. 
For sampled frames, we compute energy maps for the LL, LH, HL, and HH bands after 2D DWT, and measure their Pearson correlations with the spatial saliency map. 
As shown in Fig.~\ref{fig:band_saliency_analysis}, the LL band exhibits relatively weak correlation with the saliency
map, while the LH, HL, and HH bands show consistently stronger
positive correlations.
This empirical association motivates applying query-conditioned modulation to the high-frequency detail bands while leaving the low-frequency approximation unchanged. 

\section{Conclusion}
In this paper, we present \ours, a training-free framework that formulates video token condensation for Large Vision-Language Models from the spatial pixel domain to the joint signal-frequency domain.
By using DWT to separate low-frequency structure from high-frequency details, \ours stabilizes temporal budget allocation and preserves query-relevant spatial evidence under aggressive compression. 
Extensive experiments on long-form video benchmarks show that \ours consistently outperforms recent token compression methods, retaining up to 99.6\% of the full-input performance under a 10\(\times\) compression ratio. 
We hope this frequency-oriented perspective provides a useful direction for multimodal video understanding.


{
    \small
    \bibliographystyle{ieeenat_fullname}
    \bibliography{main}

@article{wang2025internvl3,
  title={Internvl3. 5: Advancing open-source multimodal models in versatility, reasoning, and efficiency},
  author={Wang, Weiyun and Gao, Zhangwei and Gu, Lixin and Pu, Hengjun and Cui, Long and Wei, Xingguang and Liu, Zhaoyang and Jing, Linglin and Ye, Shenglong and Shao, Jie and others},
  journal={arXiv preprint arXiv:2508.18265},
  year={2025}
}

@article{bai2025qwen2,
  title={Qwen2.5-vl technical report},
  author={Bai, Shuai and Chen, Keqin and Liu, Xuejing and Wang, Jialin and Ge, Wenbin and Song, Sibo and Dang, Kai and Wang, Peng and Wang, Shijie and Tang, Jun and others},
  journal={arXiv preprint arXiv:2502.13923},
  year={2025}
}

@article{zhang2024video,
  title={Video instruction tuning with synthetic data},
  author={Zhang, Yuanhan and Wu, Jinming and Li, Wei and Li, Bo and Ma, Zejun and Liu, Ziwei and Li, Chunyuan},
  journal={arXiv preprint arXiv:2410.02713},
  year={2024}
}

@article{zhang2023video,
  title={Video-llama: An instruction-tuned audio-visual language model for video understanding},
  author={Zhang, Hang and Li, Xin and Bing, Lidong},
  journal={arXiv preprint arXiv:2306.02858},
  year={2023}
}

@article{li2023videochat,
  title={Videochat: Chat-centric video understanding},
  author={Li, KunChang and He, Yinan and Wang, Yi and Li, Yizhuo and Wang, Wenhai and Luo, Ping and Wang, Yali and Wang, Limin and Qiao, Yu},
  journal={arXiv preprint arXiv:2305.06355},
  year={2023}
}

@article{hurst2024gpt,
  title={Gpt-4o system card},
  author={Hurst, Aaron and Lerer, Adam and Goucher, Adam P and Perelman, Adam and Ramesh, Aditya and Clark, Aidan and Ostrow, AJ and Welihinda, Akila and Hayes, Alan and Radford, Alec and others},
  journal={arXiv preprint arXiv:2410.21276},
  year={2024}
}

@inproceedings{lin2024video,
  title={Video-llava: Learning united visual representation by alignment before projection},
  author={Lin, Bin and Ye, Yang and Zhu, Bin and Cui, Jiaxi and Ning, Munan and Jin, Peng and Yuan, Li},
  booktitle={Proceedings of the 2024 Conference on Empirical Methods in Natural Language Processing},
  pages={5971--5984},
  year={2024}
}

@article{li2024llava,
  title={Llava-onevision: Easy visual task transfer},
  author={Li, Bo and Zhang, Yuanhan and Guo, Dong and Zhang, Renrui and Li, Feng and Zhang, Hao and Zhang, Kaichen and Zhang, Peiyuan and Li, Yanwei and Liu, Ziwei and others},
  journal={arXiv preprint arXiv:2408.03326},
  year={2024}
}

@inproceedings{luo2026quota,
  title={Quota: Query-oriented token assignment via cot query decouple for long video comprehension},
  author={Luo, Yongdong and Chen, Wang and Huang, Weizhong and Yin, Shukang and Lin, Haojia and Huang, Jinfa and Fu, Chaoyou and Ji, Jiayi and Zheng, Xiawu and Luo, Jiebo},
  booktitle={Proceedings of the AAAI Conference on Artificial Intelligence},
  volume={40},
  number={29},
  pages={24160--24168},
  year={2026}
}

@article{liu2023visual,
  title={Visual instruction tuning},
  author={Liu, Haotian and Li, Chunyuan and Wu, Qingyang and Lee, Yong Jae},
  journal={Advances in neural information processing systems},
  volume={36},
  pages={34892--34916},
  year={2023}
}

@article{li2022blip,
  title={BLIP: Bootstrapping Language-Image Pre-training for Unified Vision-Language Understanding and Generation},
  author={Li, Junnan and Li, Dongxu and Xiong, Caiming and Hoi, Steven},
  journal={arXiv preprint arXiv:2201.12086},
  year={2022}
}

@inproceedings{li2023blip,
  title={Blip-2: Bootstrapping language-image pre-training with frozen image encoders and large language models},
  author={Li, Junnan and Li, Dongxu and Savarese, Silvio and Hoi, Steven},
  booktitle={International conference on machine learning},
  pages={19730--19742},
  year={2023},
  organization={PMLR}
}

@inproceedings{li2024llama,
  title={Llama-vid: An image is worth 2 tokens in large language models},
  author={Li, Yanwei and Wang, Chengyao and Jia, Jiaya},
  booktitle={European Conference on Computer Vision},
  pages={323--340},
  year={2024},
  organization={Springer}
}

@inproceedings{liu2024st,
  title={St-llm: Large language models are effective temporal learners},
  author={Liu, Ruyang and Li, Chen and Tang, Haoran and Ge, Yixiao and Shan, Ying and Li, Ge},
  booktitle={European Conference on Computer Vision},
  pages={1--18},
  year={2024},
  organization={Springer}
}

@inproceedings{chen2024image,
  title={An image is worth 1/2 tokens after layer 2: Plug-and-play inference acceleration for large vision-language models},
  author={Chen, Liang and Zhao, Haozhe and Liu, Tianyu and Bai, Shuai and Lin, Junyang and Zhou, Chang and Chang, Baobao},
  booktitle={European Conference on Computer Vision},
  pages={19--35},
  year={2024},
  organization={Springer}
}

@inproceedings{yang2025vflowopt,
  title={Vflowopt: A token pruning framework for lmms with visual information flow-guided optimization},
  author={Yang, Sihan and Xu, Runsen and Cui, Chenhang and Wang, Tai and Lin, Dahua and Pang, Jiangmiao},
  booktitle={Proceedings of the IEEE/CVF International Conference on Computer Vision},
  pages={23924--23934},
  year={2025}
}

@inproceedings{shang2025llava,
  title={Llava-prumerge: Adaptive token reduction for efficient large multimodal models},
  author={Shang, Yuzhang and Cai, Mu and Xu, Bingxin and Lee, Yong Jae and Yan, Yan},
  booktitle={Proceedings of the IEEE/CVF International Conference on Computer Vision},
  pages={22857--22867},
  year={2025}
}

@article{xing2024pyramiddrop,
  title={Pyramiddrop: Accelerating your large vision-language models via pyramid visual redundancy reduction},
  author={Xing, Long and Huang, Qidong and Dong, Xiaoyi and Lu, Jiajie and Zhang, Pan and Zang, Yuhang and Cao, Yuhang and He, Conghui and Wang, Jiaqi and Wu, Feng and others},
  journal={arXiv preprint arXiv:2410.17247},
  year={2024}
}

@article{zeng2025glimpse,
  title={A glimpse to compress: Dynamic visual token pruning for large vision-language models},
  author={Zeng, Quan-Sheng and Li, Yunheng and Wang, Qilong and Jiang, Peng-Tao and Wu, Zuxuan and Cheng, Ming-Ming and Hou, Qibin},
  journal={arXiv preprint arXiv:2508.01548},
  year={2025}
}

@article{liu2025hiprune,
  title={HiPrune: Training-Free Visual Token Pruning via Hierarchical Attention in Vision-Language Models},
  author={Liu, Jizhihui and Du, Feiyi and Zhu, Guangdao and Lian, Niu and Li, Jun and Chen, Bin},
  journal={arXiv preprint arXiv:2508.00553},
  year={2025}
}

@article{zhang2025beyond,
  title={Beyond Attention or Similarity: Maximizing Conditional Diversity for Token Pruning in MLLMs},
  author={Zhang, Qizhe and Liu, Mengzhen and Li, Lichen and Lu, Ming and Zhang, Yuan and Pan, Junwen and She, Qi and Zhang, Shanghang},
  journal={arXiv preprint arXiv:2506.10967},
  year={2025}
}

@article{huang2025tosa,
  title={ToSA: Token Merging with Spatial Awareness},
  author={Huang, Hsiang-Wei and Chai, Wenhao and Chen, Kuang-Ming and Yang, Cheng-Yen and Hwang, Jenq-Neng},
  journal={arXiv preprint arXiv:2506.20066},
  year={2025}
}

@article{bolya2022token,
  title={Token merging: Your vit but faster},
  author={Bolya, Daniel and Fu, Cheng-Yang and Dai, Xiaoliang and Zhang, Peizhao and Feichtenhofer, Christoph and Hoffman, Judy},
  journal={arXiv preprint arXiv:2210.09461},
  year={2022}
}

@article{shen2025fastvid,
  title={Fastvid: Dynamic density pruning for fast video large language models},
  author={Shen, Leqi and Gong, Guoqiang and He, Tao and Zhang, Yifeng and Liu, Pengzhang and Zhao, Sicheng and Ding, Guiguang},
  journal={arXiv preprint arXiv:2503.11187},
  year={2025}
}

@article{shen2024tempme,
  title={Tempme: Video temporal token merging for efficient text-video retrieval},
  author={Shen, Leqi and Hao, Tianxiang and He, Tao and Zhao, Sicheng and Zhang, Yifeng and Liu, Pengzhang and Bao, Yongjun and Ding, Guiguang},
  journal={arXiv preprint arXiv:2409.01156},
  year={2024}
}

@article{shao2025holitom,
  title={HoliTom: Holistic Token Merging for Fast Video Large Language Models},
  author={Shao, Kele and Tao, Keda and Qin, Can and You, Haoxuan and Sui, Yang and Wang, Huan},
  journal={arXiv preprint arXiv:2505.21334},
  year={2025}
}

@inproceedings{yang2025visionzip,
  title={Visionzip: Longer is better but not necessary in vision language models},
  author={Yang, Senqiao and Chen, Yukang and Tian, Zhuotao and Wang, Chengyao and Li, Jingyao and Yu, Bei and Jia, Jiaya},
  booktitle={Proceedings of the Computer Vision and Pattern Recognition Conference},
  pages={19792--19802},
  year={2025}
}

@inproceedings{tao2025dycoke,
  title={DyCoke: Dynamic Compression of Tokens for Fast Video Large Language Models},
  author={Tao, Keda and Qin, Can and You, Haoxuan and Sui, Yang and Wang, Huan},
  booktitle={Proceedings of the Computer Vision and Pattern Recognition Conference},
  pages={18992--19001},
  year={2025}
}

@article{fan2026flashvid,
  title={Flashvid: Efficient video large language models via training-free tree-based spatiotemporal token merging},
  author={Fan, Ziyang and Chen, Keyu and Xing, Ruilong and Li, Yulin and Jiang, Li and Tian, Zhuotao},
  journal={arXiv preprint arXiv:2602.08024},
  year={2026}
}

@inproceedings{wan-etal-2024-look,
    title = "{LOOK}-{M}: Look-Once Optimization in {KV} Cache for Efficient Multimodal Long-Context Inference",
    author = "Wan, Zhongwei  and
      Wu, Ziang  and
      Liu, Che  and
      Huang, Jinfa  and
      Zhu, Zhihong  and
      Jin, Peng  and
      Wang, Longyue  and
      Yuan, Li",
    editor = "Al-Onaizan, Yaser  and
      Bansal, Mohit  and
      Chen, Yun-Nung",
    booktitle = "Findings of the Association for Computational Linguistics: EMNLP 2024",
    month = nov,
    year = "2024",
    address = "Miami, Florida, USA",
    publisher = "Association for Computational Linguistics",
    url = "https://aclanthology.org/2024.findings-emnlp.235/",
    doi = "10.18653/v1/2024.findings-emnlp.235",
    pages = "4065--4078"
}

@inproceedings{yao2022wave,
  title={Wave-vit: Unifying wavelet and transformers for visual representation learning},
  author={Yao, Ting and Pan, Yingwei and Li, Yehao and Ngo, Chong-Wah and Mei, Tao},
  booktitle={European conference on computer vision},
  pages={328--345},
  year={2022},
  organization={Springer}
}

@article{wang2025fourier,
  title={Fourier-vlm: Compressing vision tokens in the frequency domain for large vision-language models},
  author={Wang, Huanyu and Kai, Jushi and Bai, Haoli and Hou, Lu and Jiang, Bo and He, Ziwei and Lin, Zhouhan},
  journal={arXiv preprint arXiv:2508.06038},
  year={2025}
}

@article{mangalam2023egoschema,
  title={Egoschema: A diagnostic benchmark for very long-form video language understanding},
  author={Mangalam, Karttikeya and Akshulakov, Raiymbek and Malik, Jitendra},
  journal={Advances in Neural Information Processing Systems},
  volume={36},
  pages={46212--46244},
  year={2023}
}

@article{wu2024longvideobench,
  title={Longvideobench: A benchmark for long-context interleaved video-language understanding},
  author={Wu, Haoning and Li, Dongxu and Chen, Bei and Li, Junnan},
  journal={Advances in Neural Information Processing Systems},
  volume={37},
  pages={28828--28857},
  year={2024}
}

@inproceedings{fu2025video,
  title={Video-mme: The first-ever comprehensive evaluation benchmark of multi-modal llms in video analysis},
  author={Fu, Chaoyou and Dai, Yuhan and Luo, Yongdong and Li, Lei and Ren, Shuhuai and Zhang, Renrui and Wang, Zihan and Zhou, Chenyu and Shen, Yunhang and Zhang, Mengdan and others},
  booktitle={Proceedings of the Computer Vision and Pattern Recognition Conference},
  pages={24108--24118},
  year={2025}
}

@inproceedings{wang2025lvbench,
  title={LVBench: An Extreme Long Video Understanding Benchmark},
  author={Wang, Weihan and He, Zehai and Hong, Wenyi and Cheng, Yean and Zhang, Xiaohan and Qi, Ji and Gu, Xiaotao and Huang, Shiyu and Xu, Bin and Dong, Yuxiao and Ding, Ming and Tang, Jie},
  booktitle={Proceedings of the IEEE/CVF International Conference on Computer Vision},
  year={2025}
}
}

\clearpage
\appendix

\section*{Appendix}
\addcontentsline{toc}{section}{Appendix}

\noindent\textbf{Appendix Contents}
\vspace{0.4em}

\begingroup
\small
\setlength{\parindent}{0pt}
\setlength{\parskip}{2pt}
\hyperref[app:experimental-protocol]{A\quad Supplemental Experimental Protocol}\\
\hyperref[app:controlled-wavelet-ablations]{B\quad Controlled Wavelet Ablations}\\
\hyperref[app:long-video-generalization]{C\quad Additional Long-Video Generalization Results}\\
\hyperref[app:feature-distribution-wsm]{D\quad Feature Distribution Analysis of WSM}\\
\hyperref[app:query-type-analysis]{E\quad Query-Type Analysis on VideoMME}\\
\hyperref[app:runtime-complexity]{F\quad Runtime and Complexity Breakdown}\\

\endgroup

\vspace{0.8em}
\hrule
\vspace{1.0em}
\section{Supplemental Experimental Protocol}
\label{app:experimental-protocol}

We evaluate WaveZip on three video LVLM backbones, including LLaVA-Video, LLaVA-OneVision, and Qwen2.5-VL. Unless otherwise specified, all experiments use 64 uniformly sampled frames per video and are conducted on NVIDIA A800 GPUs with 80GB memory. The per-frame visual token budget is set to 196, which follows the default visual-token configuration used by LLaVA-Video and LLaVA-OneVision and falls within the valid visual-token range of Qwen2.5-VL. For each retained-token ratio \(\rho\), the total visual-token budget is defined as \(B=\rho N_v\), where \(N_v\) denotes the number of original visual tokens before compression.
For efficiency evaluation, we report end-to-end inference latency and peak GPU memory. Latency is measured \textbf{under a single-video, single-query setting} and includes the cross-modal scorer, WaveZip compression, and LVLM inference. We \textbf{do not parallelize} the scorer and LVLM inference, and we \textbf{do not use cross-query feature caching} in the reported results. This protocol provides a conservative measurement of the full WaveZip pipeline and avoids hiding the overhead introduced by the auxiliary scorer.
Each sample is evaluated once under deterministic decoding, and the
reported latency and peak memory are averaged over the fixed
evaluation subset.

\section{Controlled Wavelet Ablations}
\label{app:controlled-wavelet-ablations}
To isolate the effect of the two wavelet-based components, we conduct a controlled mechanism ablation on VideoMME with LLaVA-Video. 
We vary the temporal allocation strategy and the spatial modulation strategy while keeping the same cross-modal scorer, retained-token budget, backbone, and evaluation subset. For temporal allocation, Raw Top-\(k\) directly allocates tokens according to the original frame-level relevance scores, while WTA applies wavelet-based temporal relevance rectification. For spatial modulation, Saliency-Only uses the spatial saliency map without wavelet decomposition, while WSM applies saliency-guided modulation to the high-frequency wavelet subbands.

\begin{figure}[t]
    \centering
    
    \includegraphics[width=1.0\linewidth]{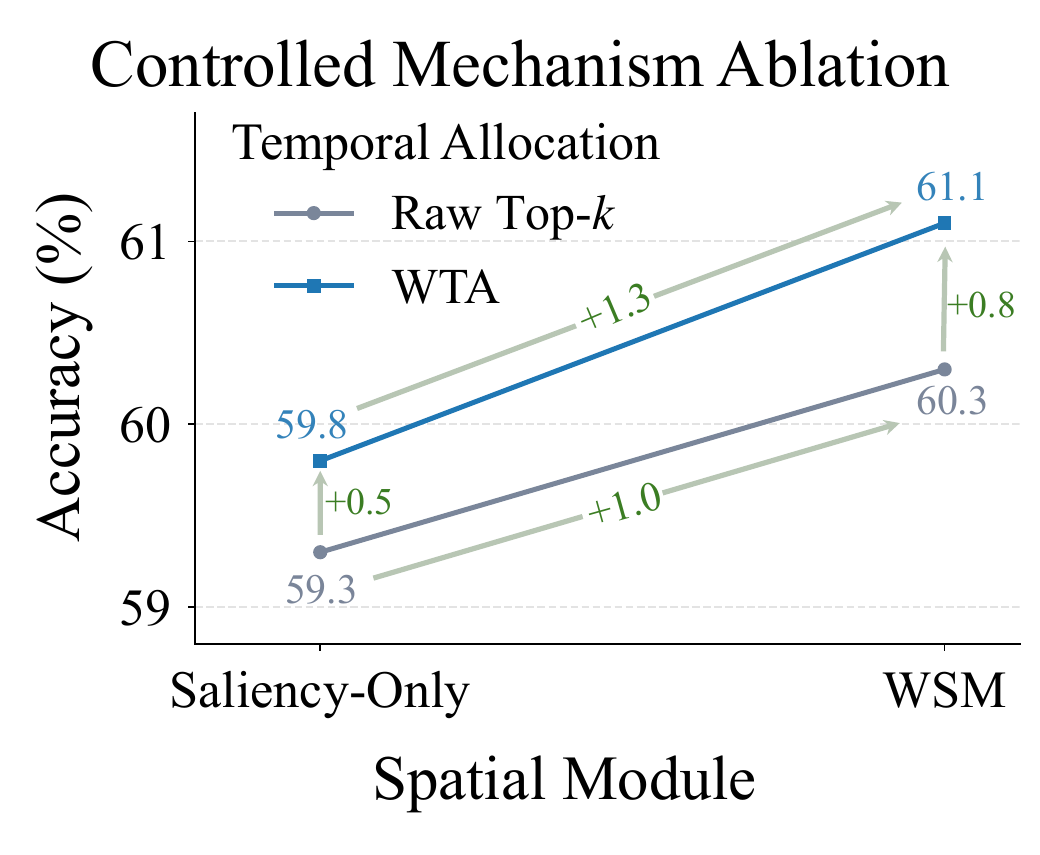}
    \caption{\textbf{Controlled mechanism ablation on VideoMMEwith LLaVA-Video-7B.}
    We compare Raw Top-\(k\) vs. WTA for temporal allocation and Saliency-Only vs. WSM for spatial modulation under the same scorer, token budget, backbone, and evaluation subset.
    All configurations use 64 sampled frames and \(\rho=0.1\).
    }
    \label{fig:controlled_mechanism_ablation}
\end{figure}

\begin{figure}[!t]
    \centering
    
    \includegraphics[width=1.0\linewidth]{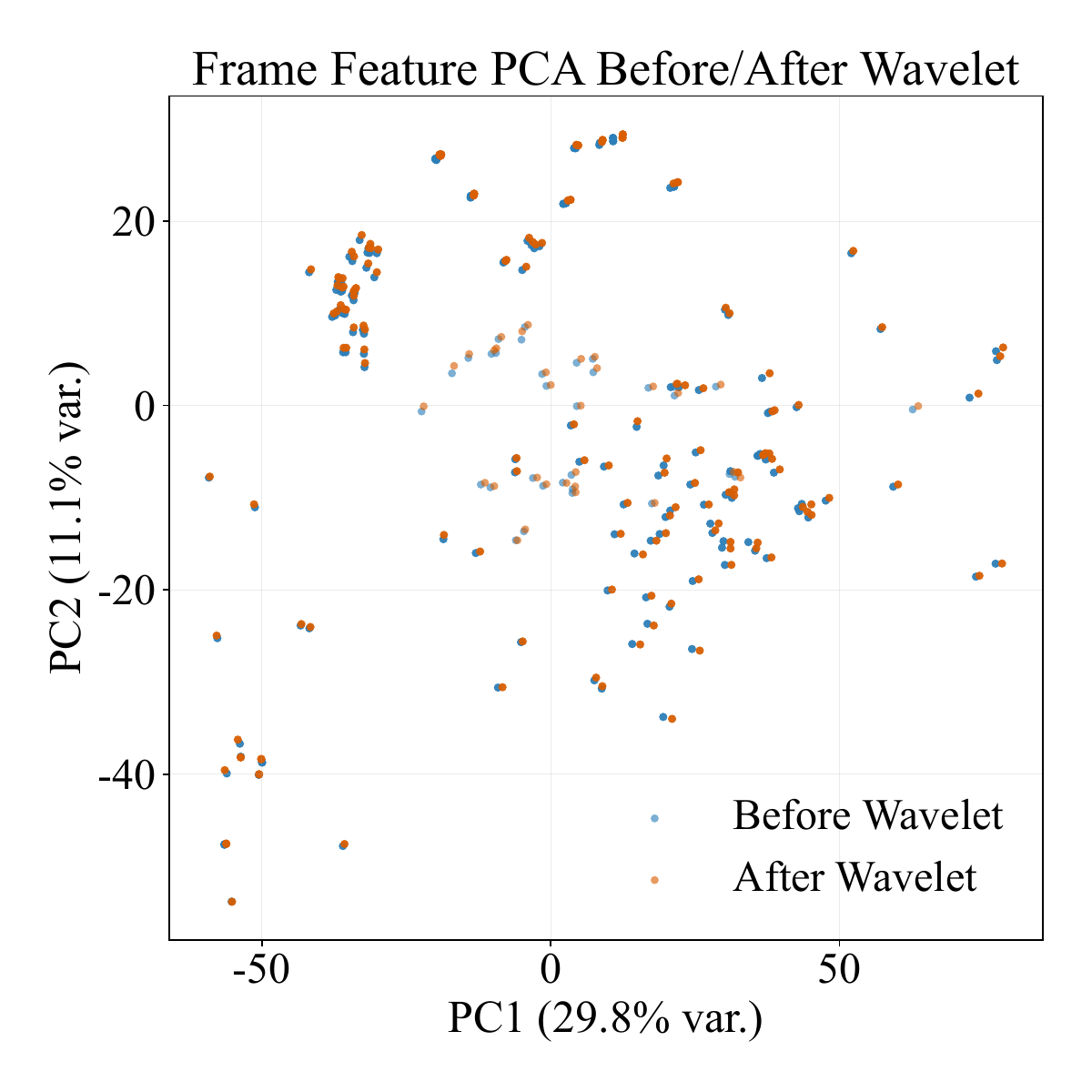}
    \caption{\textbf{PCA visualization of frame-level features before and after WSM.}
    We project the original features and the WSM-modulated features into the same PCA space. The first two principal components explain 29.8\% and 11.1\% of the variance, respectively.}
    \label{fig:feature_pca_wsm}
\end{figure}

As shown in Fig.~\ref{fig:controlled_mechanism_ablation}, WTA improves over Raw Top-\(k\) under both spatial settings, increasing accuracy from \(59.3\%\) to \(59.8\%\) with Saliency-Only and from \(60.3\%\) to \(61.1\%\) with WSM. This indicates that the temporal gain does not come from naive relevance ranking, but from wavelet-based relevance rectification. Similarly, WSM improves over Saliency-Only under both temporal allocation settings, increasing accuracy from \(59.3\%\) to \(60.3\%\) with Raw Top-\(k\) and from \(59.8\%\) to \(61.1\%\) with WTA. This confirms that the spatial gain comes from frequency-aware high-frequency modulation rather than plain saliency weighting. The best result is achieved when WTA and WSM are combined, supporting their complementary roles in temporal budget allocation and spatial detail preservation.

\begin{table}[t]
\centering
\small
\setlength{\tabcolsep}{7pt}
\renewcommand{\arraystretch}{1.0}
\begin{tabular}{c l c}
\toprule
\makecell{Retained\\Ratio $\rho$} & Method & LVBench (\%) $\uparrow$ \\
\midrule

100\%
& \cellcolor{gray!20}{Vanilla}
& \cellcolor{gray!20}{41.8} \\

\midrule

\multirow{3}{*}{20\%}
& VisionZip & 36.5 \\
& FastVID   & 39.0 \\
& \cellcolor{cyan!10}{\textbf{WaveZip}}
& \cellcolor{cyan!10}{\textbf{41.2}} \\

\midrule

\multirow{3}{*}{10\%}
& VisionZip & 31.4 \\
& FastVID   & 38.5 \\
& \cellcolor{cyan!10}{\textbf{WaveZip}}
& \cellcolor{cyan!10}{\textbf{40.9}} \\

\bottomrule
\end{tabular}
\caption{\textbf{Generalization to LVBench with LLaVA-Video-7B.}
Vanilla uses the full visual tokens, while the remaining methods are evaluated under two retained-token ratios.
All scores are accuracy (\%).}
\label{tab:app-lvbench}
\end{table}
\begin{figure}[t]
    \centering
    \includegraphics[width=\linewidth]{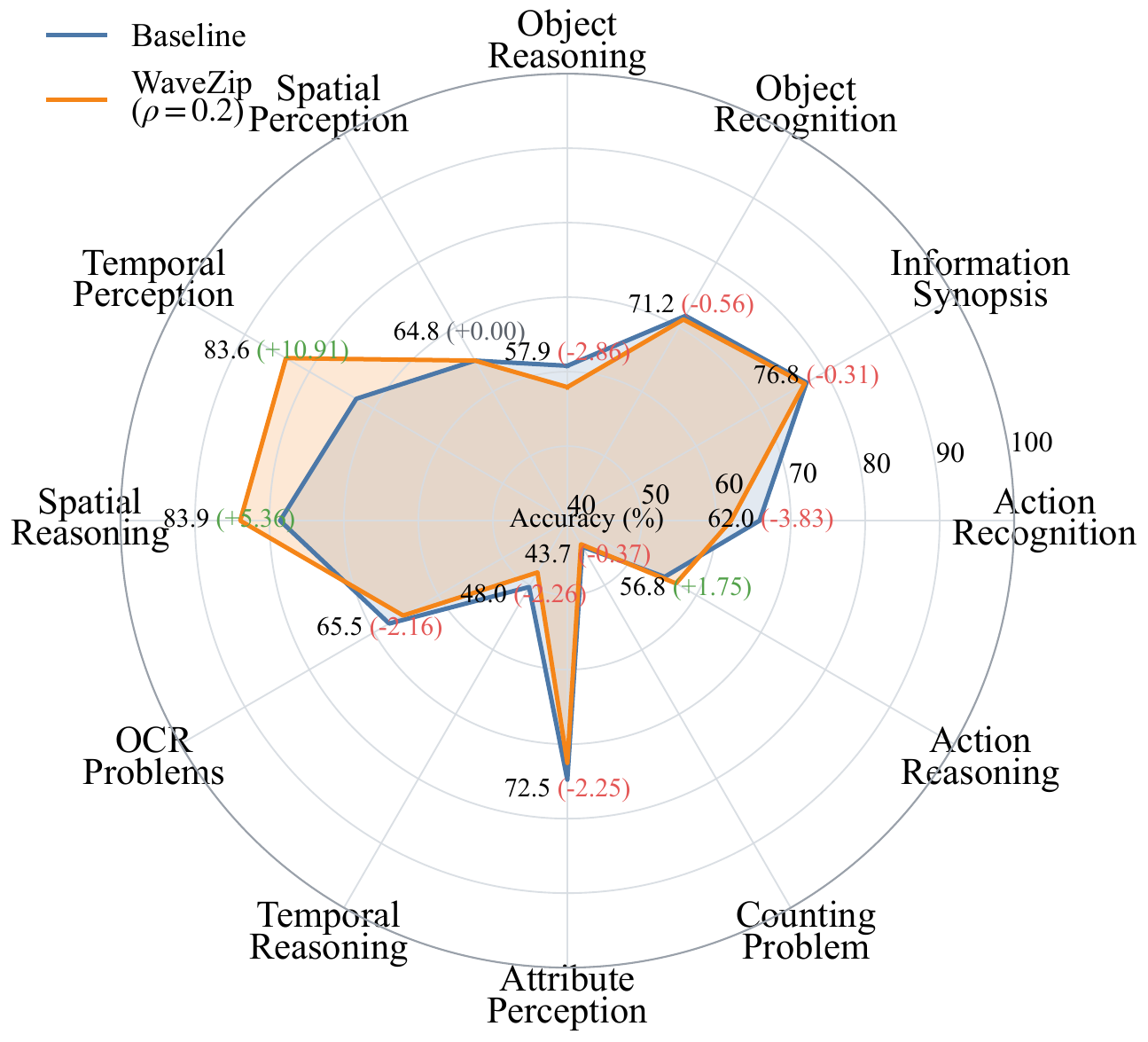}
    \caption{\textbf{Query-type analysis on VideoMME.}
    We compare the full-input baseline and WaveZip at \(\rho=0.2\). The number shown for each category is the WaveZip accuracy, and the value in parentheses denotes the difference relative to the baseline.}
    \label{fig:query_type_analysis}
\end{figure}
\section{Additional Long-Video Generalization Results}
\label{app:long-video-generalization}
We further evaluate WaveZip on LVBench~\cite{wang2025lvbench}, an extreme long-video understanding benchmark, to examine whether the method generalizes beyond the three benchmarks used in the main evaluation.
As shown in Table~\ref{tab:app-lvbench}, WaveZip consistently outperforms prior compression baselines at both retained-token ratios.
At \(\rho=0.2\), it retains 98.6\% of the full-input accuracy and exceeds the strongest compression baseline by 2.2 points; at the more aggressive \(\rho=0.1\), it remains within 0.9 points of the full-input result and surpasses the strongest baseline by 2.4 points.
These results provide additional evidence that WaveZip transfers to extreme long-video understanding beyond the three benchmarks used in the main evaluation.

\section{Feature Distribution Analysis of WSM}
\label{app:feature-distribution-wsm}
To examine whether the wavelet-based spatial modulation (WSM) introduces undesirable feature drift in the training-free setting, we analyze the distribution of frame-level visual features before and after WSM. Specifically, we project both the original features and the WSM-modulated features into a shared two-dimensional PCA space, and further summarize the distributional change using several quantitative statistics.

As shown in Fig.~\ref{fig:feature_pca_wsm}, the feature distribution after WSM remains highly consistent with that before modulation. The two point clouds exhibit strong overlap, and most paired samples stay close to their original positions without noticeable cluster migration or large-scale geometric distortion. This suggests that WSM does not alter the global feature organization of the backbone, but instead performs localized adjustments around the original representation.

\begin{table}[!t]
\centering
\begin{tabular}{lc}
\toprule
Metric & Relative Change \\
\midrule
Centroid shift / original radius & 2.7\% \\
Paired L2 change / feature norm & 3.5\% \\
Covariance trace change & 1.5\% \\
\bottomrule
\end{tabular}
\caption{\textbf{Quantitative feature-shift statistics before and after WSM.}}
\label{tab:feature_shift_stats}
\end{table}

\begin{table}[!t]
\centering

\begin{tabular}{lccc}
\toprule
Component & Time (s) & Share (\%) & Complexity \\
\midrule
WTA & 0.03 & 1.3 & \(\mathcal{O}(N)\) \\
WSM & 0.01 & 0.4 & \(\mathcal{O}(N)\) \\
Cross-modal scorer & 0.35 & 15.6 & \(\mathcal{O}(N^2)\) \\
LVLM encoding & 1.38 & 61.3 & \(\mathcal{O}((\rho N)^2)\) \\
LVLM generation & 0.48 & 21.3 & \(\mathcal{O}(L_{\mathrm{gen}}\rho N)\) \\
\midrule
Total & 2.25 & 100.0 & -- \\
\bottomrule
\end{tabular}
\caption{\textbf{Per-sample runtime and complexity breakdown of WaveZip.}
Measurements are conducted on VideoMME with LLaVA-Video using 64-frame sampling and 20\% token retention. Latency is measured on a single NVIDIA A800 GPU and an Intel(R) Platinum 8350C CPU with 32 cores.}
\label{tab:runtime_complexity_breakdown}
\end{table}

Table~\ref{tab:feature_shift_stats} further confirms this observation. 
The centroid shift is only 2.7\% relative to the original feature radius, the average paired feature change is 3.5\% of the feature norm, and the covariance trace changes by only 1.5\%. 
Together, these results indicate that WSM introduces only a limited global distributional shift while modifying individual frame representations. 
Combined with the downstream accuracy results, this suggests that WSM remains compatible with the pretrained LVLM feature space in the training-free setting.

\section{Query-Type Analysis on VideoMME}
\label{app:query-type-analysis}
To better understand how WaveZip behaves under different question types, we report a category-wise breakdown on VideoMME. We compare the full-input baseline with WaveZip at \(\rho=0.2\), and visualize the per-category accuracy in Fig.~\ref{fig:query_type_analysis}. The value next to each category denotes the accuracy of WaveZip, while the value in parentheses shows the difference relative to the baseline.

As shown in Fig.~\ref{fig:query_type_analysis}, WaveZip yields the largest gains on
Temporal Perception ($+10.91 points$), Spatial Reasoning ($+5.36$), and Action Reasoning ($+1.75$), while matching the full-token result on Spatial Perception ($+0.00$). 
Moderate reductions are observed on Information Synopsis ($-0.31$), Object Recognition ($-0.56$), Counting Problem ($-0.37$), OCR Problems ($-2.16$), Attribute Perception ($-2.25$), and Temporal Reasoning ($-2.26$). 
The largest drops occur on Action Recognition ($-3.83$) and Object Reasoning ($-2.86$).
These category-level results show that the effect of token
compression varies across question types. A more detailed
causal analysis would require controlled evaluation according
to evidence duration and spatial granularity.

\section{Runtime and Complexity Breakdown}
\label{app:runtime-complexity}
We provide a component-level runtime breakdown of WaveZip to clarify the computational overhead introduced by different modules. The measurement is conducted on VideoMME with LLaVA-Video using 64-frame sampling and 20\% token retention. The reported latency is measured under the same single-video, single-query setting as the efficiency evaluation, without parallelizing the cross-modal scorer and LVLM inference.

As shown in Table~\ref{tab:runtime_complexity_breakdown}, WTA and WSM introduce only minor overhead, jointly accounting for 1.7\% of the total latency.
The main computational cost comes from LVLM encoding and generation, while the cross-modal scorer contributes a moderate but non-negligible overhead.
Since the reported efficiency numbers include the scorer and do not use parallel execution or feature caching, this breakdown reports the full serial execution cost of the WaveZip pipeline without hiding the overhead of the auxiliary scorer.

\end{document}